# Human vs. Computer Go: Review and Prospect

Chang-Shing Lee[*], Mei-Hui Wang
Department of Computer Science and Information Engineering, National University of Tainan, *TAIWAN*

Shi-Jim Yen
Department of Computer Science and Information Engineering, National Dong Hwa University, *TAIWAN*

Ting-Han Wei, I-Chen Wu
Department of Computer Science, National Chiao Tung University, *TAIWAN*

Ping-Chiang Chou, Chun-Hsun Chou
Taiwan Go Association, *TAIWAN*

Ming-Wan Wang
Nihon Ki-in Go Institute, *JAPAN*

Tai-Hsiung Yang
Haifong Weiqi Academy, *TAIWAN*

**Abstract**

The Google DeepMind challenge match in March 2016 was a historic achievement for computer Go development. This article discusses the development of computational intelligence (CI) and its relative strength in comparison with human intelligence for the game of Go. We first summarize the milestones achieved for computer Go from 1998 to 2016. Then, the computer Go programs that have participated in previous IEEE CIS competitions as well as methods and techniques used in AlphaGo are briefly introduced. Commentaries from three high-level professional Go players on the five AlphaGo versus Lee Sedol games are also included. We conclude that AlphaGo beating Lee Sedol is a huge achievement in artificial intelligence (AI) based largely on CI methods. In the future, powerful computer Go programs such as AlphaGo are expected to be instrumental in promoting Go education and AI real-world applications.

## I. Computer Go Competitions

The IEEE Computational Intelligence Society (CIS) has funded human vs. computer Go competitions in IEEE CIS flagship conferences since 2009. Fig. 1 shows the year and the location of the conferences. The descriptions of competitions held from 1998 to 2016 are listed in detail in an online version of this article [1-8]. The handicaps for the human vs. computer 19×19 game have been decreased from 29 in 1998 to 0 in 2016. The skill of amateur players in Go is ranked according to *kyu* (K) in the lower tier, where a smaller number stands for stronger playing skill (with 1K being the highest skill level), and *dan* (D) in the higher tier, where a larger number stands for stronger playing skill. Professional Go players are ranked entirely in dan, abbreviated with the letter *P* (e.g. Lee Sedol is ranked at 9P). In the amateur level, each difference in rank roughly translates to a single stone of *handicap* (H), where the weaker player is allowed to place an additional stone on the board prior to play to even out the game. The skill difference between professional ranks is much less than one stone for every rank difference. Go is typically played on 19×19 size boards, but 9×9 size boards are also common for beginners. The complexity of the 9×9 game is far less than the standard game, and the 9×9 game had been one of the interim goals for computer Go programs. Go is a game that is inherently biased for the first player to play, Black. To compensate for this first player advantage, White is awarded additional points at the end of the game, which is referred to as *komi*. The related statistics for the IEEE CIS human vs. computer Go competitions are listed in the online version of this article [8]. It is worth noting that with a komi of 7.5, White may end up with an advantage in both 9×9 games and handicapped 19×19 games, regardless of whether White is played by humans or computers. Fig. 2 shows the certificate awarded to the MoGoTW program (16cores / 48GB / 9×9) by the Taiwan Go Association in 2010 for playing at a 3D amateur level

---

[*] Corresponding author: Chang-Shing Lee (E-mail: leecs@mail.nutn.edu.tw).

[1]. In this article, we attempt to demonstrate the massive barrier that competing programs need to overcome to achieve comparable performance to AlphaGo. For more information, eleven programs (Aya, CGI, Coldmilk/Jimmy, Erica, Fuego, MoGo/MoGoTW, Many Faces of Go, Pachi, and Zen) from 7 countries that have participated in past IEEE CIS conferences are listed in alphabetical order in the online version of this article [8].

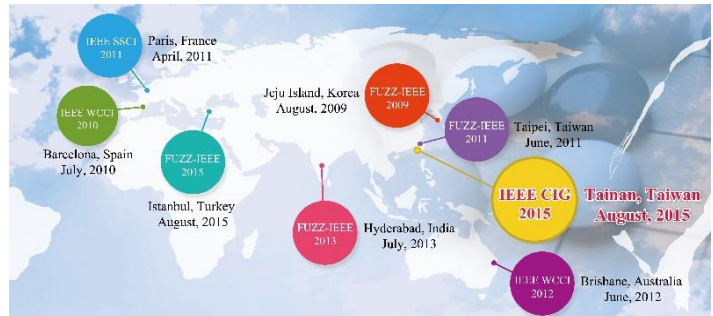
Fig. 1. Past human vs. computer Go competitions in IEEE CIS flagship conferences.

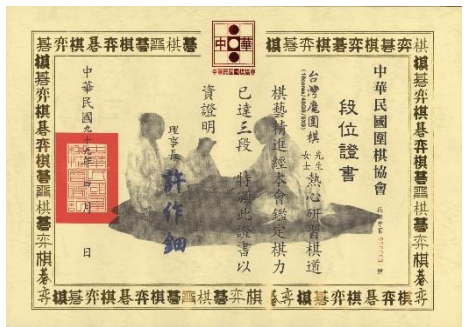
Fig. 2. Amateur 3D certificate awarded to MoGoTW in 2010.

## II. AlphaGo

In this section, we briefly introduce the past techniques used in computer Go programs, then provide an estimate for why AlphaGo is able to outperform contemporary programs so dramatically. Currently, Monte Carlo tree search (MCTS), minorization-maximization (MM), and deep convolutional neural networks (DCNNs) have demonstrated great success in Go. MCTS was successfully applied to Go in 2006 [9, 10], leading to a significant improvement in playing skill. One year later, MM was applied so that programs may recognize move patterns using supervised learning, with expert game records as the training sample [11]. Though not as revolutionary as MCTS, MM has also had a long-lasting impact on Go programs from 2007 up to 2014. In December of 2014, two teams (one of which is the DeepMind AlphaGo team) applied DCNNs to Go independently [12, 13]. Clark and Storkey [12] first published a paper that applied DCNNs to Go, which, when given a game position, could estimate how expert human players respond with a prediction rate of 41%-44%, exceeding the rate of previous methods. Meanwhile, DeepMind's method, which was released 10 days later, had a prediction rate of 55%. Among many of DCNN's applications, it has seen success in image and video recognition. When applied to Go, DCNN is able to recognize move patterns at a significantly lower error rate than MM. For this reason, most state-of-the-art computer Go programs use MCTS combined with either MM or DCNN.

AlphaGo is able to perform leaps and bounds above other contemporary programs because of its extensive use of high quality neural networks, which cannot be easily reproduced by other teams due to insufficient experience and/or inadequate hardware resources. To illustrate, let us consider the three main neural networks used in AlphaGo: a supervised learning (SL) policy network, a reinforcement learning (RL) policy network, and the value network [14]. Both the SL policy network and the value network were used in AlphaGo during competitive play; the RL policy network was used only for generating the training samples for the value network. The SL policy network takes a game board position and attempts to guess where expert players will play next. This SL process was performed with 30 million game positions, and involved 340 million training steps, taking a total of three weeks with 50 graphic processing units (GPUs) [14]. The SL algorithm tends to mimic what it has learned from game records instead of favoring moves that yield the highest winning rate when given a choice. To improve on this, the SL policy network was used to train a separate policy network using RL. Training for the RL policy network takes one day with 50 GPUs. The key to AlphaGo's playing skill is its value network, which was trained through RL with 30 million self-play game positions. The training process takes about one week with 50 GPUs, for a total training time of four weeks and a day for all three networks.

The most time-consuming and most difficult process to reproduce, however, is not the training of these three networks, but the generation of self-play game positions. For each of the 30 million self-play positions, 100 playouts are performed; for each playout, we assume on average 200 moves until game completion, so a total of 600 billion move data samples need to be generated to train the value network. Let us assume, for the sake of demonstration, that a research team has access to four GPUs. The training of the three networks will take [(4 weeks × 7 days / week) + 1 day] × $\frac{50 \text{ GPUs}}{4 \text{ GPUs}}$ = 362.5 days. Assuming a processing speed of 720 moves/s for a single GPU (with a batch size of 16), an optimistic estimate for the generation of self-play game samples is $600 \times 10^9$ moves / (4 × 720 moves / second) $\cong$ 208 million seconds, which works out to 2411 days or about 80 months. In addition to the total time required for generating and training the networks, we must consider the fact that parameters involved in the entire process are rarely tuned to fit the requirements in a single trial. This includes a wide variety of settings such as the number of layers and neurons in the neural network, the features to use for the Go positions, the collection of expert game records that are used to train the initial SL policy network, etc. This quick estimate of required resources does not even take into account the knowledge and experience that the DeepMind team has acquired since its inception. As a side note, the distributed version of AlphaGo uses 280 GPUs [14].

## III. Human Intelligence View

Demis Hassabis, CEO and Co-Founder of Google DeepMind, described Go as the "Mt. Everest" of AI [15] because Go is a very complex board game that requires intuitive, creative, and strategic thinking [16]. In the past decade, the techniques of MCTS had revolutionized the field of computer game-playing. The playing strength of computer Go programs has progressed to about a four-stone handicap against top professional Go players in 2012-2015. More precisely, until AlphaGo's emergence in Oct. 2015, the world's strongest program, Zen, was able to beat a top professional Go player with a handicap of four stones, while losing when the handicap was decreased to three stones. Google DeepMind introduced a new approach that combined MCTS with deep learning in their program AlphaGo [14], which subsequently broke this four-stone handicap barrier in the recent competition with Lee Sedol (9P) in Korea, in Mar. 2016. AlphaGo's performance sent shock waves through the community of professional Go players and AI researchers. Since then, people on the Internet and media, particularly in Go-playing cultures such as in Korea, Taiwan, Japan, and China, were buzzing with discussion on related topics [17-19]. In this section, we invited three high-level professional Go players who have spent time helping the development of computer Go programs, including Coldmilk and MoGoTW, and who have also been part of shaping Go trends for many years, to comment on the game results of the challenge match from Mar. 9 to Mar. 15, 2016. Fig. 3 shows the game record for match No. 5 and Comment 1 lists three professional Go players' brief commentaries. For the other four matches' commentaries and the full commentary on match No. 5, readers can refer to the online version of this article [8]. Additionally, readers may find details for the Go terminologies used in Comments 1 and 2 at Sensei's Library [20]. From the perspective of professional Go player (Ping-Chiang Chou / 6P), the strengths and weaknesses of AlphaGo are listed in Comment 2.

Comment 1.    Professional Go players' comments on match No. 5.

| |
|---|
| **Comments by Ping-Chiang Chou (6P / Taiwan)** |
| Theoretically, Black made a profit from the fight in the bottom right corner. Hence, up to White 68, Black's situation was slightly better from the view point of professional players. Chou concluded that perhaps Black 79 was decided under the premise that Black already had the lead. If the match was against other professional players under the same situation, Black 79 would have intuitively been played at L14. In reality, from White 80 to White 90 (jump), the situation had turned favorable for White. |
| **Comments by Chun-Hsun Chou (9P / Taiwan)** |
| White started to play an unusual move at move 18. White 20, 22, and 24 are indicative of AlphaGo's unconventional playing style, which sacrifices a few stones to gain a sente and obtain the maximum profit. Black 79 and 81 were two slow moves intended for stability. Unfortunately, they were shut in by White 80 and 82 so the situation favored White by a little. Although there were variations in the following moves, White took hold of the situation with a weak superiority until the endgame. As a result, Black ended up losing by about 2.5 points after komi 7.5. |
| **Comments by Ming-Wan Wang (9P / Japan)** |
| Shortly after the game began, White again made mistakes, allowing Black to have the upper hand. This unexpected turn of events caused Black (Lee Sedol) to change his playing style so that he may conservatively aim to make life. After being laid siege by White, Black realized he was behind, and started to make every effort to regain his lead, whereupon the situation became quite chaotic. It was unfortunate that Black did not play the strongest move at the critical moment, which ended the game with a small loss for Black. |

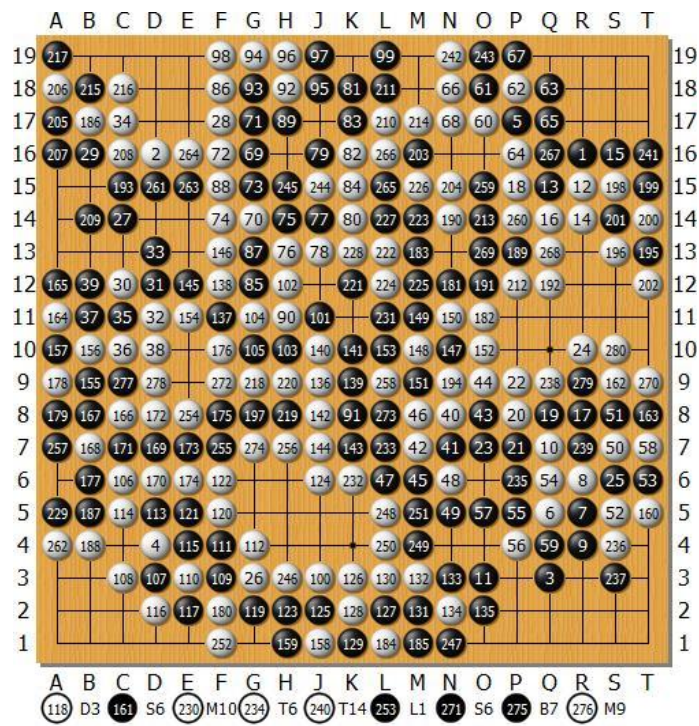

Fig. 3.  Game record (Black is Lee Sedol, White is AlphaGo, and White won by resignation) for match No. 5 on Mar. 15, 2016.

Comment 2.    Overall comments from human Go players.

**Overall Comments by Ming-Wan Wang (9P / Japan)**
- Playing Go is not just a matter of achieving victory, but the knowledge involved and enjoyment of the game are also important. Currently, it seems to be difficult for top professional Go players to beat AlphaGo. However, the purpose of playing against human Go players is not simply for the victory itself, but also for communicating and sharing the game-playing process with each other. Go is an interesting game that leads to a lot of profound discoveries. Sometimes, we pay too much attention to improving our strength in playing skill, neglecting other things which we should focus on. We should explore Go from another perspective to enhance our enjoyment of the game.
- Each side (White or Black) needs to play at least over 100 moves during each Go game. Humans might lose the game from only one mistake in a decisive battle. In addition to the disadvantages in physical and mental conditions, humans do not adopt strategies from a whole-game perspective. A human Go player could manage to find a critical move to increase the winning rate, but fail to evaluate all of the possible moves. Nevertheless, an inevitable result is a human Go player's subjective thoughts, which perhaps influence him / her to correctly analyze game situations.

**Overall Comments by Ping-Chiang Chou (6P / Taiwan)**
- AlphaGo's strengths: AlphaGo shows very strong judgement both for the whole board and for local situations. Additionally, AlphaGo is also able to accurately calculate territory / points.
- AlphaGo's weaknesses: When the situation favors AlphaGo, its playing style will become conservative. AlphaGo is capable of strong local computation but it does not seem to choose the most optimal move to play, for example, the fight in the bottom right corner on match No. 5 with Lee Sedol. It tends to avoid making kos if there are other lines of thinking to choose from.

**Overall Comments by Tai-Hsiung Yang (7D / Taiwan)**
- The aim of the computer Go program is to pursue the maximum probability to win instead of winning the maximum amount of territory. That is why AlphaGo sometimes loses some points by playing irrational moves when it apparently takes the lead. Human Go players suffer through terrible psychological stress when playing with such a strong computer Go program as AlphaGo. On the other hand, AlphaGo can also help find blind points that are easily missed by humans.
- It is not easy to differentiate between the middle game and the endgame. Sometimes, the endgame will start after the fighting. Other times, the fighting and the endgame will simultaneously happen. The computer Go program's computational speed is several times more powerful than the human Go player's. Accordingly, developing a software for Go training is absolutely useful for strengthening professional Go players' endgame skill.
- The game of Go contains a lot of the joy and philosophy of life; it is not just a game concerned with winning, but a cultural legacy that cultivates human talent.

## IV. Discussions and Future Studies

It has been estimated that when human players play against a new computer program for the first time, their strengths are usually weakened by about 1-2 levels. This is because human players will usually play a probe move when they have no idea about the program's strength. Another key point is that Go programs tend to be designed differently from how human Go players tend to think; winning ten points is essentially the same as winning one point as far as the program's objective function is concerned. As a result, the Go programs tend to maximize the winning rates, while human players tend to maximize territory. Additionally, programs have other advantages when playing against humans, such as relatively strong raw computational power and the lack of facial expressions, all of which in contrast are human disadvantages. There are weaknesses in computer programs, however, such as hidden bugs, imperfect judgement on the overall situation at the beginning of the game, and imperfect parameter adjustment. According to cognitive psychologists [21], the difference between experts (for example, Go teachers or professional Go players) and newcomers is that the former can store a variety of Go patterns in their brains, allowing them to quickly find a more precise move than the latter. Upon seeing the formed shapes of stones on the board, they are able to scan their memory to find the best matching pattern, then play an effective move. Newcomers, on the other hand, are short of such a quick connection.

Computer Go, in particular AlphaGo, has been an extremely hot topic lately, not only to AI researchers but also to the general public. The purpose of this article is to help the readership better understand how the development of computer Go programs has arrived at this milestone of winning against one of the top human players, and how CIS has been involved in this process. This article's main contribution is to place the win of AlphaGo over Lee Sedol in recent historical context. This huge achievement in AI is based largely on CI methods, including DCNNs, SL from expert games, RL, the use of the value network and policy network, and MCTS. The strength of AlphaGo, especially as measured against the other computer Go programs, is absolutely amazing. Playing with contemporary computer Go programs like Crazy Stone, Zen, Pachi, Fuego, and GnuGo, with no handicaps on either side, the single-machine version of AlphaGo was able to win 494 games out of 495 in total, while the distributed version of AlphaGo won all games against these competing programs [14].

As an alternative method of evaluation, the DeepMind team used an Elo rating scale, where each player gets a numerical strength estimation computed from past game results. AlphaGo, human European champion Fan Hui, and the above listed competing programs were evaluated. The distributed version of AlphaGo had an Elo rating of 3140, the non-distributed version was at 2890, and Fan Hui was at 2908 [14]. Since then, AlphaGo's playing strength has grown even stronger. At the time of writing, according to an online ranking [22] curated by Go programmer and author of Crazy Stone, Rémi Coulom, distributed AlphaGo's Elo rating is 3590 (the second highest rating in the world), where Ke Jie (3624) and Lee Sedol (3525) are ranked as the first and fourth highest Elo rating in the world, respectively. It is worth noting that the Elo rating system computed in [22] is not exactly the same as the one used in [14]. The ratings are given here only for the benefit of establishing some intuition for AlphaGo's progress. In addition to the professional players' comments on the five games between AlphaGo and Lee Sedol, this article also provides overall comments from Ming-Wan Wang (9P / Japan), Ping-Chiang Chou (6P / Taiwan), and Tai-Hsiung Yang (7D / Taiwan, director of Haifong Weiqi Academy, Taiwan) in Comment 2. AlphaGo's victory over the world champion Lee Sedol in Mar. 2016 will be marked in history as a remarkable achievement. However, this would not have been possible without the considerable time and effort of countless contributors to computer Go in the past. AlphaGo's impact will almost assuredly popularize and improve Go learning worldwide, especially if a personalized version with reduced hardware costs becomes available. Finally, AlphaGo's performance was truly astonishing, and will undoubtedly be a continued source of inspiration for professional Go players and AI researchers around the world.

## Acknowledgements

The authors would like to thank the Ministry of Science and Technology of Taiwan for its financial support under the grant MOST 105-2919-I-024-001-A1, MOST 104-2221-E-024-015, and MOST 104-2622-E-024-005-CC2. Additionally, the authors also would like to thank 1) Mr. Ti-Rong Wu (National Chiao Tung University, Taiwan) for providing the resource estimate in generating self-play game positions (see Section II); 2) Mr. Sheng-Shu Chang (President of Click108 Company, Taiwan) for his financial support on past human vs. computer Go competitions @ IEEE CIS flagship conferences; and 3) Dr. Olivier Teytaud and INRIA TAO team members as well as Taiwanese team members and National Center for High-Performance Computing (NCHC) under the grant NSC 99-2923-E-024-003-MY3. Finally, we would like to thank the anonymous referees for their constructive and useful comments.

## References

[1] M.-H. Wang, C.-S. Lee, Y.-L. Wang, M.-C. Cheng, O. Teytaud, and S.-J. Yen, "The 2010 contest: MoGoTW vs. human Go players," *ICGA Journal*, vol. 33, no. 1, pp. 47–50, Mar. 2010.

[2] C.-S. Lee, M.-H. Wang, G. Chaslot, J.-B. Hoock, A. Rimmel, O. Teytaud, S.-R. Tsai, S.-C. Hsu, and T.-P. Hong, "The computational intelligence of MoGo revealed in Taiwan's computer Go tournaments," *IEEE Transactions on Computational Intelligence and AI in Games*, vol. 1, no. 1, pp. 73–89, Mar. 2009.

[3] A. Rimmel, O. Teytaud, C.-S. Lee, S.-J. Yen, M.-H. Wang, and S.-R. Tsai, "Current frontiers in computer Go," *IEEE Transactions on Computational Intelligence and AI in Games*, vol. 2, no. 4, pp. 229–238. Dec. 2010.

[4] C.-S. Lee, M.-H. Wang, O. Teytaud, and Y.-L. Wang, "The game of Go @ IEEE WCCI 2010," *IEEE*

*Computational Intelligence Magazine*, vol. 5, no. 4, pp. 6–7, Nov. 2010.

[5] C.-S. Lee, O. Teytaud, M.-H. Wang, and S.-J. Yen, "Computational intelligence meets game of Go @ IEEE WCCI 2012," *IEEE Computational Intelligence Magazine*, vol. 7, no. 4, pp. 10–12, Nov. 2012.
[6] C.-S. Lee, M.-H. Wang, M.-J. Wu, O. Teytaud, and S.-J. Yen, "T2FS-based adaptive linguistic assessment system for semantic analysis and human performance evaluation on game of Go," *IEEE Transactions on Fuzzy Systems*, vol. 23, no. 2, pp. 400–420, Apr. 2015.
[7] J.-B. Hoock, C.-S. Lee, A. Rimmel, F. Teytaud, M.-H. Wang, and O. Teytaud, "Intelligent agents for the game of Go," *IEEE Computational Intelligence Magazine*, vol. 5, no. 4, pp. 28–42, Nov. 2010.
[8] C.-S. Lee, M.-H. Wang, S.-J. Yen, T.-H. Wei, I.-C. Wu, P.-C. Chou, C.-H. Chou, M.-W. Wang, and T.-H. Yang, "Human vs. computer Go: review and prospect @ IEEE CIS: integrated human and computational intelligence for future Go learning based on Google AlphaGo's historic achievement," Apr. 2016, [Online] Available: https://sites.google.com/site/nutnoaselabenglish/computergo.
[9] R. Coulom, "Efficient selectivity and backup operators in Monte-Carlo tree search," in H. Jaap van den Herik, P. Ciancarini, and H. H. L. M. (Jeroen) Donkers (editors), Computers and Games, Berlin, Heidelberg, Springer, 2006, pp. 72–83.
[10] L. Kocsis and C. Szepesvári, "Bandit based Monte-Carlo planning," in Proceeding of 17th European Conference on Machine Learning (ECML 2006), Berlin, Germany, Sept. 18–22, 2006, pp. 282–293.
[11] R. Coulom, "Computing Elo ratings of move patterns in the game of Go," in Proceeding of Computer Games Workshop 2007 (CGW 2007), Amsterdam, The Netherlands, Jun. 15–17, 2007, pp. 113–124.
[12] C. Clark and A. Storkey, "Teaching deep convolutional neural networks to play Go," Dec. 2014, [Online] Available: http://arxiv.org/abs/1412.3409.
[13] C. J. Maddison, A. Huang, I. Sutskever, and D. Silver, "Move evaluation in Go using deep convolutional neural networks," Dec. 2014, [Online] Available: http://arxiv.org/abs/1412.6564.
[14] D. Silver, A. Huang, C. J. Maddison, A. Guez, L. Sifre, G. van den Driessche, J. Schrittwieser, I. Antonoglou, V. Panneershelvam, M. Lanctot, S. Dieleman, D. Grewe, J. Nham, N. Kalchbrenner, I. Sutskever, T. Lillicrap, M. Leach, K. Kavukcuoglu, T. Graepel, and D. Hassabis, "Mastering the game of Go with deep neural networks and tree search," *Nature*, vol. 529, pp. 484–489, Jan. 2016.
[15] Taipei Times, "Human go champ beats supercomputer," Mar. 14, 2016, [Online] Available: http://www.taipeitimes.com/News/front/archives/2016/03/14/2003641528.
[16] Wikipedia, "AlphaGo versus Lee Sedol," Mar. 2016, [Online] Available: https://en.wikipedia.org/wiki/AlphaGo_versus_Lee_Sedol#Difficult_challenge_in_artificial_intelligence.
[17] Wired, "Google's AI wins first game in historic match with Go champion," Mar. 2016, [Online] Available: http://www.wired.com/2016/03/googles-ai-wins-first-game-historic-match-go-champion/?mbid=nl_3916.
[18] Wired, "In two moves, AlphaGo and Lee Sedol redefined the future," Mar. 2016, [Online] Available: http://www.wired.com/2016/03/two-moves-alphago-lee-sedol-redefined-future/.
[19] D. Hassabis, "Official Google Blog: What we learned in Seoul with AlphaGo," Mar. 2016 [Online] Available: https://googleblog.blogspot.tw/2016/03/what-we-learned-in-seoul-with-alphago.html?m=1.
[20] A. Hollosi and M. Pahle, "Sensei's library," Apr. 2016, [Online] Available: http://senseis.xmp.net/.
[21] F. Gobet and M. H. Ereku, "Computer program beats European Go champion," Feb. 2016, [Online] Available: https://www.psychologytoday.com/blog/inside-expertise/201602/computer-program-beats-european-go-champion.
[22] R. Coulom, "Go Ratings," Apr. 2016, [Online] Available: http://www.goratings.org/.